# SentiGOLD: A Large Bangla Gold Standard Multi-Domain Sentiment Analysis Dataset and its Evaluation

Md. Ekramul Islam
Giga Tech Limited
Dhaka, Bangladesh
ekramul.islam@gigatechltd.com

Labib Chowdhury
Giga Tech Limited
Dhaka, Bangladesh
labib.chowdhury@gigatechltd.com

Faisal Ahamed Khan
Giga Tech Limited
Dhaka, Bangladesh
faisal.cse06@gigatechltd.com

Shazzad Hossain
Giga Tech Limited
Dhaka, Bangladesh
shazzad.hossain@gigatechltd.com

Sourave Hossain
Giga Tech Limited
Dhaka, Bangladesh
sourave.hossain@gigatechltd.com

Mohammad Mamun Or Rashid
Bangladesh Computer Council
Dhaka, Bangladesh
mamunbd@juniv.edu

Nabeel Mohammed
North South University
Dhaka, Bangladesh
nabeel.mohammed@northsouth.edu

Mohammad Ruhul Amin
Fordham University
New York, USA
mamin17@fordham.edu

## ABSTRACT

In this study, we present a Bangla multi-domain sentiment analysis dataset, named as SentiGOLD, developed using 70,000 samples, which was compiled from a variety of sources and annotated by a gender-balanced team of linguists. This dataset was created in accordance with a standard set of linguistic conventions that were established after multiple meetings between the Government of Bangladesh and a nationally recognized Bangla linguistics committee. Although there are standard sentiment analysis datasets available for English and other rich languages, there are not any such datasets in Bangla, especially because, there was no standard linguistics framework agreed upon by national stakeholders. SentiGOLD derives its raw data from online video comments, social media posts and comments, blog posts and comments, news and numerous other sources. Throughout the development of this dataset, domain distribution and class distribution were rigorously maintained. SentiGOLD was created using data from a total of 30 domains (e.g. politics, entertainment, sports, etc.) and was labeled using 5 classes (e.g. strongly negative, weakly negative, neutral, weakly positive, and strongly positive). In order to maintain annotation quality, the national linguistics committee approved an annotation scheme to ensure a rigorous Inter Annotator Agreement (IAA) in a multi-annotator annotation scenario. This procedure yielded an IAA score of 0.88 using Fleiss' kappa method, which is elaborated upon in the paper. A protocol for intra- and cross-dataset evaluation was utilized in our efforts to develop a classification system as a standard. The cross-dataset evaluation was performed on the SentNoB dataset, which contains noisy Bangla text samples, thereby establishing a demanding test scenario. We also performed cross-dataset testing by employing zero-shot experiments, and our best model produced competitive performance, which exemplify our dataset's generalizability. Our top model attained a macro f1 of 0.62 (intra-dataset) for 5 classes establishing the benchmark for SentiGOLD, and 0.61 (cross-dataset from SentNoB) for 3 classes which stands comparable to the current state-of-the-art. Our fine-tuned sentiment analysis model[1] can be accessed online.

## 1 INTRODUCTION

Sentiment Analysis (SA) is deemed as a standard application of Natural Language Processing (NLP) to analyze wide range of opinions from textual data. SA is primarily employed to predict the attitude embedded within a given phrase with different levels of polarity, such as positive, negative, and neutral. Positive sentiment is present in a given text if it provides a pleasant or satisfying notion. Conversely, anything is considered a negative feeling if it transmits a message of sadness, dissatisfaction, or anger. In the case of a neutral perspective, the text implies impartiality toward any entity. Oftentimes, a text having a neutral attitude is considered to be objective rather than subjective.

---
[1]https://sentiment.bangla.gov.bd

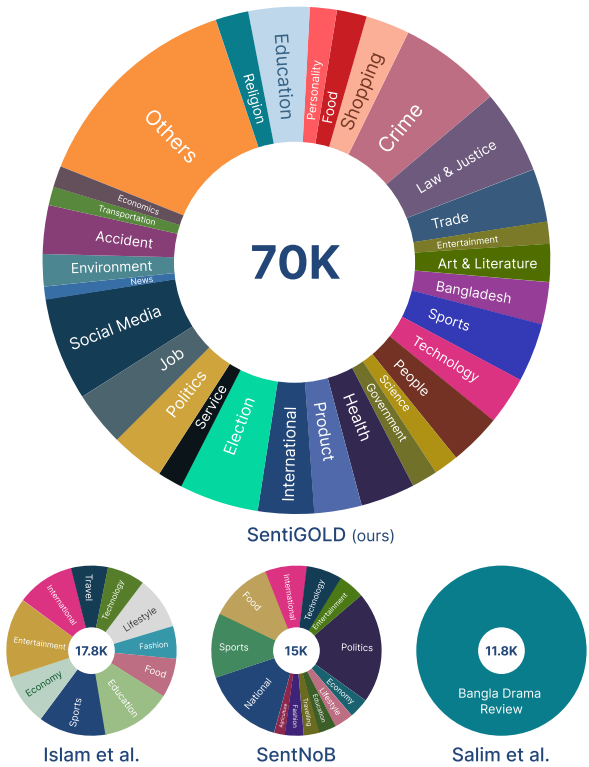

**Figure 1: Comparison between our proposed SentiGOLD with three other Bangla sentiment analysis datasets that are available online: 1. Islam et al. [17]; 2. SentNoB [19]; and 3. Salim et al. [34]. Different colors inside the pie chart indicate different domains of each dataset. Among all the datasets, SentiGOLD is the largest in terms of size and diversity.**

In recent times, SA has been increasingly used for social media insight, election campaigns, brand value monitoring, business insight, customer service, stock market predictions, and market research. SA is inherently a tricky problem to solve, as individuals may perceive the meaning within the same text quite differently, thus leading to a multitude of probable sentiment outcomes. Thus, to solve the problem with higher confidence, a finite set of different sentiments is used. And more importantly, a dataset comprising thousands of real-world texts capturing the predefined set of sentiments is required to solve the problem using NLP modeling techniques. A standard dataset must have some key characteristics. For instance, the dataset should be large and diverse enough for proper model generalization. Moreover, the dataset must have a high IAA [10] score to be reliable. To remove human bias from the dataset, blindfold annotation by multiple human annotators is also important [30]. When sourcing the dataset from different domains, a critical requirement is to maintain the class distributions to be well-balanced [40].

Even though Bangla is the sixth most spoken language in the world [16], only 0.1% of all websites include Bangla content[2]. On the other hand, languages that are used widely for communication over the Internet, such as English, Russian, Spanish, French, and German, comprise more than 75% of the top 10 million websites of the world[3]. Thus, despite some high-quality efforts made to construct a comprehensive SA dataset, a large-scale Bangla SA dataset sourced from diverse domains is needed. One of the most recent works on SA is SentNoB [19]. Although SentNoB is, in our perspective, one of the more rigorous works produced in the area of Bangla SA, it has certain shortcomings, such as - 1. a relatively small dataset with 15k samples collected from 13 domains that bring its generalizability into question; 2. the annotation of the dataset achieved a low IAA score of 0.53; and 3. most importantly, annotators of SentNoB were not formally trained in Bangla linguistics.

To mitigate those shortcomings and establish a benchmark dataset for Bangla, the Information and Communication Technology Ministry of the People's Republic of Bangladesh took the initiative to launch the project titled, "Development of Sentiment Analysis Software in Bangla (SD-18)", under Enhancement of Bangla Language in ICT through Research & Development (EBLICT) initiatives [3][14]. This project aims to create a gold-standard SA dataset of 70,000 samples and to implement a benchmark modeling for Bangla SA. The project's Terms of Reference (ToR) [15] specify the guidelines to be followed while creating the dataset and a minimum IAA score including modeling performance. Researchers from several reputed universities in Bangladesh materialized and assessed those guidelines. A team of researchers from the Bangladesh University of Engineering and Technology served as the project's testing consultant. In this article, we detail the dataset creation, its assessment, benchmarking, and other critical performance analyses compared to the previous research work, that are directly related to the successful completion of the aforementioned project.

When constructing the SentiGOLD dataset, most of the other resources we discovered had at least one significant problem. For example, SentNoB [19] contains samples with incorrectly assigned labels in the dataset. The vast majority of publications did not even measure the IAA score for their datasets [2, 12]. Another point of concern is the lack of domain diversity in those datasets (see **Figure 1**) [6, 21, 29]. It is unfortunate to report that all the previously published work failed to include professional linguists to specify the annotation scheme to construct their datasets [2, 33]. Even most of the existing SA datasets are incapable of capturing the appropriate sentiment in a real-world scenario [6, 21, 29].

To improve and mitigate the aforementioned flaws, in this paper, we present SentiGOLD: a Bangla gold standard dataset for sentiment analysis. Our main contributions to this paper are as follows:

- We propose SentiGOLD, a sentiment analysis dataset with 70,000 labeled samples. The proposed dataset has 5 classes: Strongly Negative (SN), Weakly Negative (WN), Neutral (N), Weakly Positive (WP), and Strongly Positive (SP).

---

[2]https://w3techs.com/technologies/details/cl-bn-
[3]https://en.wikipedia.org/wiki/Languages_used_on_the_Internet

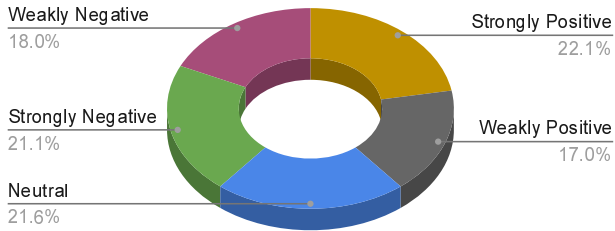

**Figure 2: Illustration of our proposed dataset's class distribution. This figure implies that SentiGOLD is well-balanced in terms of class distribution.**

- The data for SentiGOLD is collected from more than 30 different domains. To the best of our knowledge, this is the most diverse Bangla SA dataset (see **Figure 1**).
- We ensure a high IAA score of 0.88 to make the dataset as reliable as possible. We have provided detail analysis of how we achieve that score.
- We have built an annotation management system from scratch to annotate Bangla SA data. We have made both the annotation management system and SentiGOLD dataset publicly available upon request[4].
- To establish a benchmark, we have investigated different architectures and training methodologies on this dataset and achieved 0.62 macro f1 for 5 classes with BanglaBert [4].
- We employ cross-dataset testing to showcase the generalization capability of the proposed dataset. We have achieved 0.61 macro f1 score for 3 classes in the cross-dataset testing. It is also worth mentioning that previous Bangla SA work did not do cross-dataset evaluation, we are the first ones to do so.
- Moreover, We have achieved 6% performance improvement in macro f1 compared to SentNoB for 3 classes, which implies our proposed dataset has more balanced distribution than SentNoB.

## 2 RELATED WORK

The work of Minqing Hu and Bing Liu, circa 2004 on customer reviews, was the first major work done on SA as claimed in paper [2]. They proposed a *Feature-Based Opinion Mining Model*, which is now popular as the *Aspect-Based Opinion Mining* [13]. *Stanford Sentiment Treebank* [35], *Amazon Product Data*, *IMDB Movie Reviews Dataset* [26], and *Sentiment140* [11], are some benchmarking examples of sentiment analysis resources.

When it comes to data characteristics, the majority of datasets, such as [21] and [17], provided clean Bangla data after undergoing various pre-processing procedures. Certain work [12] included code-mixed samples of both Bangla and English. In order to make the model robust, some work, such as [19], produced a noisy real-world dataset that includes posts from social networks. Majority of the aforementioned research attempted two or three classes (positive, negative, and neutral) labeling. Only a few, such as [19], split both the positive and negative labels further into two separate categories based on the intensity of sentiments, such as weak or strong. A few other studies, such as [29] and [6], concentrated on widening the vocabulary size of the dataset, as enriched vocabulary set had claimed to produce better sentiment analysis models. These two research groups attempted aspect-based sentiment analysis techniques on their datasets. Cross-lingual approaches were also explored in Bangla by some researchers, such as [34] and [33], which used translation of Bangla sentences to its English counterpart for SA. This rather introduced further dependency hence did not receive much traction in the research community. On the other hand, class imbalance in the dataset of [36] restricted its efficacy as it contains much more positive or negative samples than neutral samples.

A few recent studies, such as [19] and [12], concentrated on the quality of data annotation. They employed at most 3 annotators to perform annotations of the data in order to achieve an IAA score, but the annotations were not validated by any expert linguists. Only a small number of studies [18] also validated their annotated data. Limited resources hindered large dataset production for deep learning, restricting research scope to specific domains. While [21] and [36] presented a single domain dataset, [17] contained 10 domains. However, all these aforementioned datasets suffer heavily from domain bias. Till now, SentNoB [19] dataset is considered the most robust work, which delivered the most diverse with 13 domains and a balanced dataset comprising 15,000 samples for Bangla SA. They reported an IAA [10] score of 0.53, but they have not validated their annotated data, which raises questions about the quality of the annotation.

Our work in Bangla sentiment analysis (SA) surpasses all previous efforts. Our dataset is approximately 4.5 times larger (70,000 samples) than the previous largest sentence-level dataset (15,000 samples) [19]. Each instance was blindly annotated by a group of 3 trained annotators using majority voting and validation by a validator. Disputes were resolved in follow-up meetings. This rigorous annotation process achieved a higher Inter-Annotator Agreement (IAA) score of 0.88. The dataset covers over 30 domains with a balanced class distribution. (see **Figure 2**).

## 3 DEVELOPMENT OF SENTIGOLD

### 3.1 Dataset Requirements

SentiGOLD is a multi-class classification dataset that labels each Bangla text with one of the following labels: **Strongly Positive, Weakly Positive, Neutral, Weakly Negative, and Strongly Negative**. Samples of SentiGOLD must be annotated by three human annotators and validated by one human validator, per the ToR [15]. It should include both clean and noisy formal and informal Bangla text encoded with Unicode. No duplicates are permitted. According to the ToR [15], text duplication should be handled based on the similarity score of two texts, with a suggested threshold of 0.80 (Jaccard Similarity). The specified source domains included newspaper websites, Bangla blogs, YouTube comment sections, Facebook posts and comments, technical blogs, and the Bangla wiki, among others. In order to ensure consistent and reliable labeling of the data, an annotation guideline was finalized in collaboration with linguistic experts from across the nation, as well as academics from Computer Science and Linguistics Departments [28]. The actual annotation tasks commenced online after the guideline was finalized.

---
[4]pdeblict@bcc.net.bd

## 3.2 Data collection

We gathered unprocessed text data from various sources while adhering to the criteria outlined in Section 3.1. In addition to considering the data source, it was essential to identify all of the topic-related domains prior to beginning data collection. As topic domains of interest, we listed several domains, including politics, sports, business, places, education, technology, health, religion, and entertainment, from the socioeconomic perspective of Bangladesh. **Figure 3** depicts the distribution of samples within SentiGOLD that pertain to these topics. The topic "Others" includes examples from agriculture, travel, and lifestyle, among others.

While high-level sources were identified in the ToR [15], a list of specific sources was finalized, focusing primarily on websites that receive the most traffic and contain a significant amount of public opinion in their comment sections. There are five types of sources, including social media, product reviews, news, Bangla blogs, tech blogs, and YouTube comment sections. This final list can be found in **Appendix A.1**.

After domain and source selection, a custom web crawler was developed. We developed the crawlers using the Scrapy [22] framework which is an open-source and free technology.

## 3.3 Data cleaning and processing

After collecting a sufficient amount of raw data from the sources mentioned previously, we began the data cleaning and processing steps. The data that was crawled contained various types of unnecessary noise, such as HTML tags, URLs, romanized Bangla text, repeating punctuations, repeating white spaces, other language words, non-Unicode Bangla text, and emoticons. We eliminated the majority of the noises with the exception of emoticons, as we believed they conveyed a sense of emotion. Multiple steps were taken to clean the dataset, including i) the application of Unicode normalization ii) eliminated all HTML tags and URLs from the text iii) eliminated all unnecessary white space iv) eliminated repeated punctuations v) identified the text language using TextBLOB [25] and filtered out non-Bangla data.

After sanitizing the dataset, we performed additional processing to remove duplicate samples. To identify duplicates in the dataset, we compared each sample against all other samples using both syntactic [20] and semantic similarity [23]. If both semantic and syntactic similarity between two samples was high, we eliminated one of the samples as depicted in the **Table 1**. Jaccard similarity [20] was used to measure structural similarities, while embeddings word2vec [27], doc2vec [24] was used with word mover distance (WMD) [23] to measure semantic similarity. The **Appendix A.3** contains a comprehensive elucidation of the procedure employed for eliminating duplicate data, utilizing the Jaccard similarity and WMD techniques.

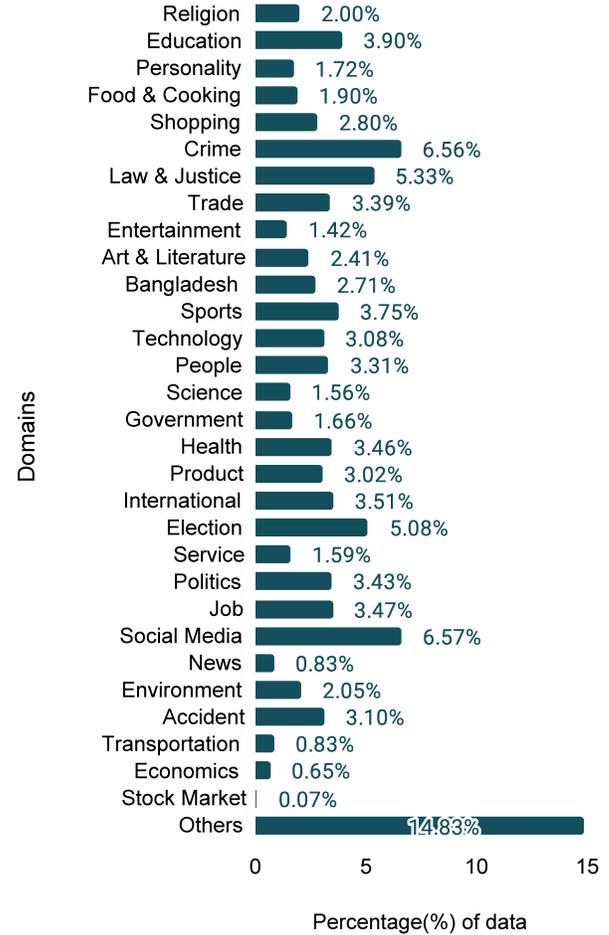

**Figure 3: Domain Distribution of topics in the proposed SentiGold dataset. "Others" tag includes topics such as agriculture, travel, and lifestyle.**

## 3.4 Annotation & Validation

As mentioned previously, an annotation guideline [28] was developed to ensure consistency of data annotation. In addition to that, an annotation procedure was developed that ensured that the data annotators and validators participation was enacted in a specified way.

The lack of readily available free and open-source annotation management systems prompted the development of a custom Annotation Management System (AMS) to support the annotation procedure. **Figure 4** depicts the entire gold-standard data annotation procedure incorporated in AMS. The AMS ensured the quality of annotated data by incorporating specific measures. A group of at least three annotators, supervised by a validator, was formed. The validator received a portion of the data assigned to the annotator group to assess their accuracy. If the annotator group's performance was below 80%, the validator discussed and identified their mistakes. Once the group achieved 80% performance, Inter-Annotator Agreement (IAA) was calculated. The final label for each instance was determined through majority voting and reviewed by the validator. The qualifications of the validators were carefully verified to establish their authority over the annotators. Detailed information about the annotators and validators is presented in **Figure 5**.

Table 1: Illustration of Duplicate Data Elimination Procedure. If both the structural and semantic similarity were high, we rejected one of the texts.

| Example | Structural Similarity | Semantic Similarity | Action |
|---|---|---|---|
| 1: এবার বর্ষায় বন্যা হতে পারে। (This time there may be flood in monsoon.) 2: বর্ষার আগে জলাবদ্ধতা দূর করতে হবে (Waterlogging should be removed before monsoon) | Low | Low | ✓ |
| 1: তিনি গান গাইতে অনেক ভালোবাসেন। (He loves to sing very much.) 2:গান না গেয়ে সে থাকতে পারে না। (He cannot live without singing.) | Low | High | ✓ |
| 1:তার গানের গলা ভালো। (His singing voice is good.) 2: তার গানের গলা ভালো না। (His singing voice is not good.) | High | Low | ✓ |
| 1: প্রধানমন্ত্রী শেখ হাসিনা গতকাল ভারত সফরে গেছেন। (Prime Minister Sheikh Hasina visited India yesterday.) 2: গতকাল শেখ হাসিনা ভারত সফরে গেছেন। (Sheikh Hasina visited India yesterday.) | High | High | ✗ |

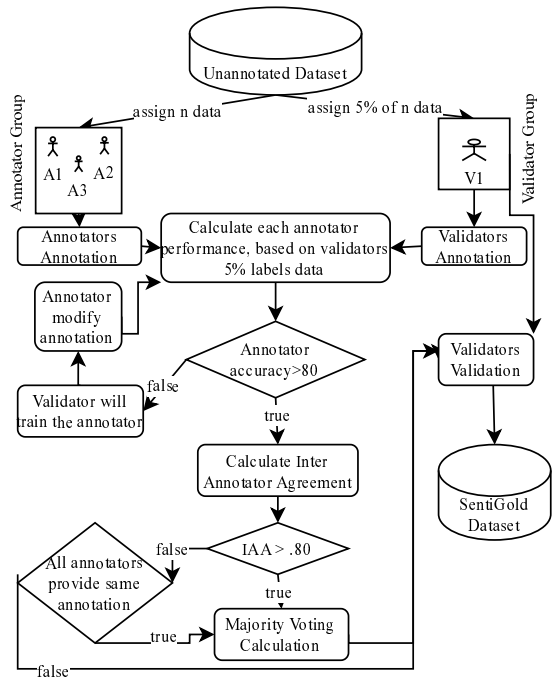

Figure 4: Illustration of our detailed process of SentiGOLD dataset development. This workflow starts by assigning unannotated data to the annotators and validators. Each sample was annotated by three linguists and validated by an expert linguist. The quality of the annotations was always monitored using the annotator's accuracy and the Inter Annotator Agreement iteratively.

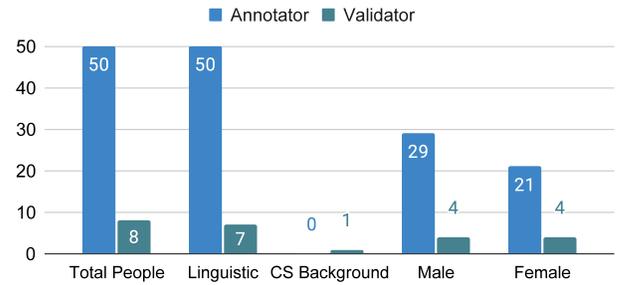

Figure 5: Bar chart of number of annotators and validators in terms of their educational background and gender.

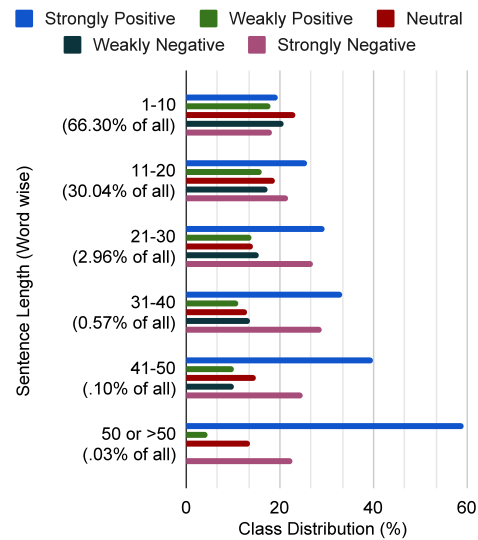

Figure 6: Illustration of Sentence Length Wise Class Distribution. The maximum length of most of our sentences is 10. We have noticed a few biases towards the Strongly Positive class when the sentence length increases.

### 3.5 Statistical Analysis

Our proposed SentiGOLD consists of 70,000 samples with an average sentence length of around ten words. As **Figure 2** illustrated, the class distribution across the whole dataset is 22.1%, 17.0%, 21.6%, 18.0% and 21.1% of the data are SP, WP, N, WN and SN respectively. **Figure 2** implies that our proposed SentiGOLD dataset is well-balanced. We further investigated our dataset in terms of sentence length-wise class distribution. From **Figure 6**, we observe a pattern that, when the sentence length increase, it tends to be a positive sentence.

The SentiGOLD dataset, developed for Bangla sentiment analysis, underwent rigorous evaluation to assess its quality using the Inter-Annotator Agreement (IAA) metric. Remarkably, SentiGOLD exhibited a consistently high IAA score of 0.88. In contrast, the SentNoB [19] dataset displayed lower IAA scores as the length of sentences increased. However, the IAA score of SentiGOLD surpassed expectations, reaching an impressive 0.98 for sentences longer than 50 characters (see **Figure 7**), indicating that longer sentences provided clearer context for sentiment annotation. To achieve such robust results, several techniques were employed. Firstly, an annotation guideline specific to Bangla sentiment analysis was meticulously crafted, and a group of 70 annotators underwent training based on trial and error. Subsequently, the 50 best performing annotators were selected, and the annotation guideline was refined based on real-life training experiences. Moreover, the annotation process involved assigning a maximum of 500 samples to each group, ensuring effective monitoring of annotator performance by designated validators. The annotation system itself facilitated real-time monitoring and training of annotators by validators, contributing to the overall accuracy of the dataset. Notably, all annotators possessed a linguistic background, and the validators were predominantly from the same department. The iterative annotation approach (see **Figure 8**) and the annotation management system's workflow were visually represented in **Figure 4**, providing insights into the methodology behind achieving a robust IAA score for the SentiGOLD dataset.

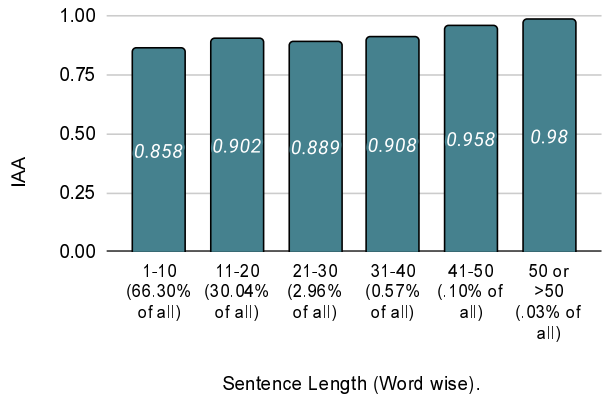

**Figure 7: Our study shows that, with appropriate annotation rules and skilled annotators, a similar pattern of Inter-Annotator Agreement (IAA) can be observed in terms of sentence length, despite the poor IAA score demonstrated by a recent Bangla sentiment analysis dataset [20].**

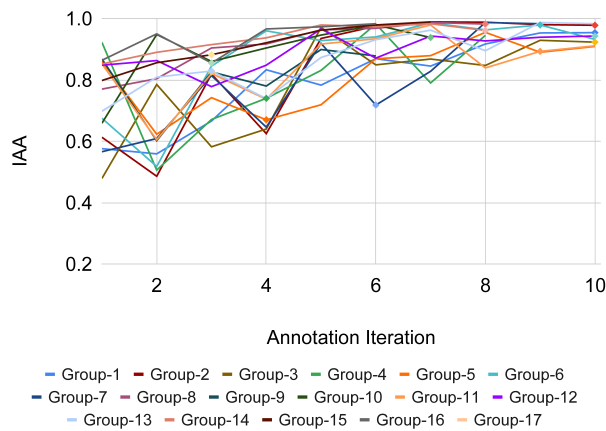

**Figure 8: We utilized 17 annotator groups, each consisting of 3 individuals, to annotate an average of 500 data samples in 10 iterations. The iterative annotation approach demonstrated gradual improvement in later iterations.**

## 4 EXPERIMENTS & ANALYSIS

### 4.1 Experimental setup

We have adopted all the experiments from [19] along with the newly released BanglaBert. We first explored some hand-crafted feature-based algorithms to know how our dataset is performing and then moved to state-of-the-art techniques such as LSTM and BERT-based algorithms. These experiments can be referred to as intra-dataset testing in the later part of the paper. To evaluate our algorithm, we conducted per-class and per-domain stratified splitting in the SentiGOLD dataset. This resulted in 80% training, 10% validation, and 10% test sets with balanced class and domain distributions. Precision, recall, and macro f1 were used as evaluation metrics, and we employed statistical tests such as the Friedman test followed by post-hoc Nemenyi test [39] to determine the rank of the models [7].

To demonstrate the generalizability of our SentiGOLD dataset, we incorporated a cross-dataset testing framework. Specifically, we compared our dataset with SentNob [19], the largest available Bangla dataset. Given that SentiGOLD is more than four times larger than SentNob, we randomly sampled 15K instances and followed the same data splitting method. This subset was referred to as SentiGOLD15K in the manuscript.

We have also performed zero shot learning approach to evaluate our dataset's generalizability. To do that We have tested our SentiGOLD finetuned BanglaBert model with [19] test set, [17] test set and the entire dataset of [34] as the author did not release any particular test set. All the experiments have been implemented using PyTorch [31], TensorFlow [1], and Transformers [37]. We have used a single Nvidia RTX 3080 to train all our models.

### 4.2 Results

We started our experiments with hand-crafted feature models such as unigram, bigram, trigram, and different combinations. We have followed the same configuration and hyper-parameter settings as [19]. After that, we moved to different combinations of BiLSTM, Hierarchical Attention Network [38] (HAN), biLSTM CNN with Attention [8] (BCA) models. After that, we moved to BERT-based pretrained language models. We adopted pretrained language models such as mBert [9] and BanglaBert [4] and finetuned with our SentiGOLD dataset.

We establish all our benchmark results in **Table 2**. We can see that with fine-tuned BanglaBert, we got our best result of 0.62 macro-f1 for 5 classes. BanglaBert is a language model trained on the raw Bangla text, which is why it worked very well while we fine-tuned it with SentiGOLD. However, we got competitive results from BCA and HAN. It is also worth mentioning that different mixing of hand-crafted feature-based algorithms performs well because we have aggregated different features from different algorithms.

Table 2: Performance evaluation for different models on 5 class SentiGOLD dataset. Finetuned BanglaBert performs better than all the configurations.

| Model | Precision | Recall | Macro F1 |
| --- | --- | --- | --- |
| Unigram | 0.47 | 0.48 | 0.47 |
| Bigram | 0.40 | 0.39 | 0.40 |
| Trigram | 0.39 | 0.28 | 0.24 |
| U+B | 0.5 | 0.5 | 0.5 |
| U+B+T | 0.5 | 0.5 | 0.49 |
| Char 2-gram | 0.45 | 0.45 | 0.44 |
| Char 3-gram | 0.48 | 0.48 | 0.48 |
| Char 4-gram | 0.48 | 0.48 | 0.48 |
| Char 5-gram | 0.50 | 0.50 | 0.50 |
| C2+C3 | 0.49 | 0.49 | 0.49 |
| C3+C4 | 0.49 | 0.49 | 0.49 |
| C4+C5 | 0.50 | 0.50 | 0.50 |
| C2+C3+C4 | 0.50 | 0.50 | 0.50 |
| C3+C4+C5 | 0.51 | 0.51 | 0.51 |
| C2+C3+C4+C5 | 0.51 | 0.51 | 0.51 |
| U+B+C3+C4+C5 | 0.50 | 0.50 | 0.50 |
| U+B+C2+C3+C4+C5 | 0.50 | 0.50 | 0.50 |
| U+B+T+C2+C3+C4+C5 | 0.51 | 0.51 | 0.51 |
| Embeddings (E) | 0.46 | 0.46 | 0.45 |
| U+B+C2+C3+C4+C5+E | 0.50 | 0.50 | 0.50 |
| U+B+T+C2+C3+C4+C5+E | 0.51 | 0.51 | 0.51 |
| Bi-LSTM + Attn. (FastText) | 0.42 | 0.41 | 0.41 |
| Bi-LSTM + Attn. (Random) | 0.44 | 0.44 | 0.43 |
| mBert | 0.38 | 0.37 | 0.37 |
| Bi-LSTM+Conv.+Attn | 0.54 | 0.53 | 0.53 |
| Hier. Attn Network | 0.53 | 0.52 | 0.52 |
| **BanglaBert** | **0.63** | **0.62** | **0.62** |

Table 3: Domain wise performance evaluation on BanglaBert. We achieved the maximum score from Job domain and the lowest was from Stock Market domain. Please note that, some samples were mapped to multiple domain.

| Domain Name | Support | Macro F1 | Micro F1 |
| --- | --- | --- | --- |
| Religion | 453 | 0.58 | 0.64 |
| Education | 588 | 0.63 | 0.63 |
| Personality | 560 | 0.61 | 0.66 |
| Food & Cooking | 258 | 0.63 | 0.63 |
| Shopping | 313 | 0.60 | 0.61 |
| Crime | 294 | 0.52 | 0.64 |
| Law and Justice | 291 | 0.56 | 0.58 |
| Trade | 162 | 0.56 | 0.57 |
| Entertainment | 451 | 0.63 | 0.65 |
| Art and Literature | 301 | 0.51 | 0.66 |
| Bangladesh | 233 | 0.61 | 0.62 |
| Sports | 242 | 0.64 | 0.63 |
| Information Technology | 284 | 0.65 | 0.63 |
| People | 288 | 0.65 | 0.68 |
| Science | 204 | 0.61 | 0.61 |
| Government | 159 | 0.58 | 0.59 |
| Health and Medicine | 338 | 0.58 | 0.58 |
| Product | 296 | 0.62 | 0.65 |
| International | 246 | 0.66 | 0.66 |
| Election | 69 | 0.59 | 0.59 |
| Service | 150 | 0.61 | 0.62 |
| Politics | 177 | 0.61 | 0.64 |
| Job | 128 | **0.69** | **0.70** |
| Social Media | 232 | 0.63 | 0.62 |
| News | 134 | 0.55 | 0.58 |
| Environment | 102 | 0.65 | 0.67 |
| Accident | 65 | 0.62 | 0.66 |
| Transportation System | 73 | 0.65 | 0.68 |
| Economics | 147 | 0.62 | 0.63 |
| Stock Market | 6 | **0.49** | **0.53** |
| Others | 1349 | 0.62 | 0.63 |

Since our proposed SentiGOLD has more than 30 domains, we further investigate how our model performs in those domains. **Table 3** represent the domain-wise macro and micro f1 score for 5 class. **Table 3** justifies our overall results as most of the macro f1 are either close to 0.62 or little more than that. We get the lowest macro f1 of 0.49 from the stock market; this is possible because 0.07% of SentiGOLD is from the stock market domain.

The statistical significance of the results was established using the Friedman test, followed by post-hoc Nemenyi test [39]. These tests compared the macro f1 scores of BanglaBert with other models across all 10-fold experiments. The results consistently showed that BanglaBert outperformed the other models significantly in terms of macro f1 score. Further details of the statistical tests can be found in **Appendix A.4**.

To further investigate, we picked random models from **Table 2** and adopted cross dataset testing framework to test the generalizability of our proposed SentiGOLD. To do that, we have trained a combination of multiple hand-crafted feature-based models, BCA, HAN, and fine-tuned BanglaBert. We have trained these models on both the SentiGOLD and SentNoB [19]. In **Table 4** also, our finetuned BanglaBert outperformed every other model with a comparatively larger margin. It is worth mentioning that SentiGOLD15K is a subset of the SentiGOLD dataset with only 15k sample to make the comparison fair with SentNoB [19]. This table also implies that our SentiGOLD dataset is much more diverse and generalized as SentiGOLD trained model always performs better than otherwise.

Table 4: Cross-dataset testing to better visualize our SentiGOLD dataset generalizability. We have trained our proposed SentiGOLD dataset with BanglaBert and tested on both three class SentNoB [19] test set and a combined curated test set from [19] and SentiGOLD. In all cases model trained on our five class SentiGOLD dataset outperformed SentNoB trained model.

| | Trained on SentNoB | | Trained on SentiGOLD | | Trained on SentiGOLD15K | |
|---|---|---|---|---|---|---|
| Model | Tested on SentiGOLD | Tested on combined | Tested on combined | Tested on SentNoB | Tested on Combined | Tested on SentNoB |
| U+B+T+C2+C3+C4+C5+E | 0.48 | 0.59 | 0.73 | 0.52 | 0.60 | 0.48 |
| Bi-LSTM+Conv.+Attn | 0.50 | 0.60 | 0.72 | 0.54 | 0.62 | 0.51 |
| Hier. Attn Network | 0.49 | 0.58 | 0.72 | 0.50 | 0.60 | 0.50 |
| BanglaBert | **0.55** | **0.70** | **0.76** | **0.61** | **0.75** | **0.61** |

Table 5: Results of zero-shot cross-dataset experiments with three other datasets of Bangla SA. Our 5 class SentiGOLD data trained model produces competitive results while testing with these datasets.

| Test Dataset | Support | Macro F1 | Micro F1 | Published Result | |
|---|---|---|---|---|---|
| | | | | Macro F1 | Micro F1 |
| SentNoB (3 Class) | 1586 | 0.61 | **0.66** | n/a | 0.64 |
| Islam et al. (3 Class) | 3000 | 0.48 | 0.53 | n/a | **0.6** |
| Salim et al. (2 Class) | 11807 | 0.9 | 0.92 | **0.93** | n/a |

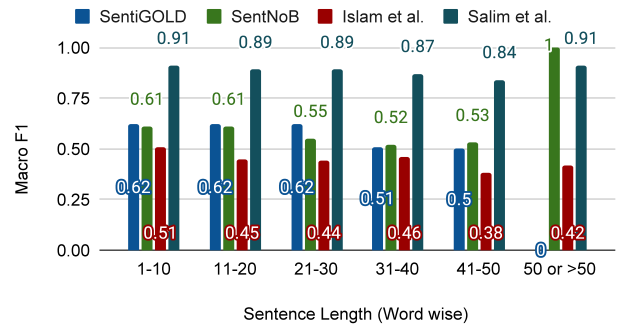

Figure 9: Our benchmark model was tested on three different datasets with variable sentence lengths, and it consistently performed well across all datasets. This evaluation was conducted using a zero-shot cross-dataset testing approach.

## 4.3 Performance analysis

To analyze our SentiGOLD dataset even further, we have adopted zero-shot learning approach to evaluate our benchmark BanglaBert model with recently proposed [19] [34] and [17] datasets. We took our SentiGOLD trained BanglaBert model and test on the aforementioned datasets. It is worth noting that, from [19] and [17] we have used their test data partition, however, in the case of [34] we have not found any partition of that dataset so we have tested their entire dataset. To perform a fair comparison, we have aggregated our strongly negative and weakly negative class to negative and strongly positive and weakly positive class to positive class. **Table 5** represent our zero-shot results, we can see that in all datasets our test results are very competitive. In case of SentNoB, our model outperformed their published result. On the other hand, Islam et al. and Salim et al. our model produces competitive results, which indicates that our proposed SentiGOLD dataset is much more diverse and generalized than the recently published dataset.

While analysing the results from our benchmark model, we have explored behavioral testing methods [32] which is a recently proposed promising procedure to test NLP models. Among the behavioral testing techniques, we have utilized the minimum functionality test [32] to assess our models. **Table 6** shows some of the behavioral test samples results.

In order to analyze how our benchmark BanglaBert model is performing on some impactful words, we have developed a test dataset by randomly collecting sentences having those words. They were collected from online resources such as Facebook, Wiki, Quora, etc. After that, a single human annotation was done by linguistic experts. **Figure 10** illustrates the results of the dataset. The lowest macro f1 score was found for the word 'Corona', which is 0.54.

To further investigate our benchmark BanglaBert model, we choose political, religious, and liberation words according to our demography. We substituted those words with synonyms or comparable words to examine how these words affect the model. For example, in **Table 7**, the difference between examples 1 and 2 is only the name "Bangabandhu" and "Mir Jafor". However, the model predicts differently because "Mir Jafor" is a name that is politically a negative word in Bangla or this subcontinent, whereas "Bangabandhu" is politically a positive word. So naturally, the bias did not come from the dataset but from the language model, it was trained on. **Figure 11** depicts the same thing, and that is why in "Mir Jafor" the count of polarity change is larger than "Bangabandhu". This implies that our proposed SentiGOLD dataset does not have any significant token bias.

Table 6: Our proposed BanglaBert successfully captures the appropriate context of the situation by comprehending the semantic meanings of phrases and the sentimental significance of individual words.

| Test Type | Test Description | Example | Expected Output | Model Output |
|---|---|---|---|---|
| Negation | Negated negative should be positive or neutral. | 1. সে একটুও অসুন্দর না। (She is not ugly at all.) | SP | SP |
| | | 2. মনে কোন দুঃখ নেই। (There is no sorrow in mind.) | SP | WP |
| | Negated neutral should still be neutral. | 1. মমতাজ সংগীত করে না। (Mumtaz doesn't do music.) | N | N |
| Semantic Role | Author sentiment is more important than of others. | 1. সবাই তোমাকে ভাল বললেও, তুমিতো আসলে ভাল না। (Everyone says you're good, but you're not really good.) | SN | SN |
| Temporal | Sentiment change overtime, present should prevail. | 1. চিপসটা আগেই মজা লাগতো কিন্তু এখন আর লাগে না। (Chips used to be fun but not anymore.) | WN | WN |
| Vocab | Sentiment should change if noun is changed from postive to negative. | 1. মূল্যবোধের মৃত্যু হয়েছে। (Values are dead.) | SN | SN |
| | | 2. দুর্নীতির মৃত্যু হয়েছে। (Corruption is dead.) | SP | SN |

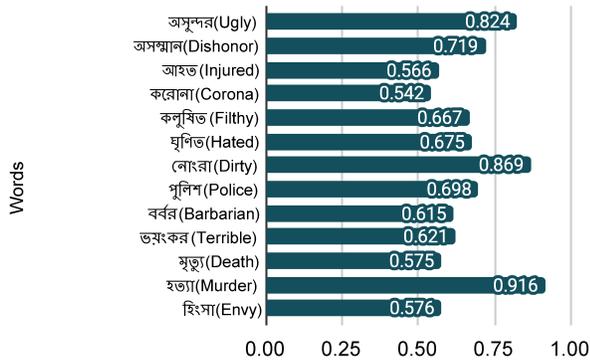

**Figure 10: Bangla is considered to be a highly inflected language [5]. Since some words itself produces negative impression according to their positioning in a sentence, we collected some socially impactful words from our linguists and performed an evaluation on BanglaBert.**

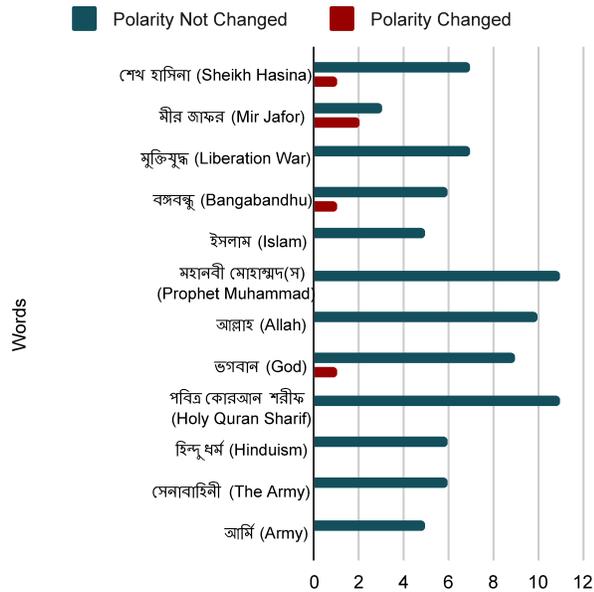

**Figure 11: These are some of the political, religious, and liberation words in Bangla Language. This figure illustrates the effectiveness in terms of sentiment polarity of these words.**

**Table 7: Illustration of token bias experiments results.**

| Text | Prediction |
| --- | --- |
| 1. বঙ্গবন্ধুর নেতৃত্ব ছাড়া মুক্তিযুদ্ধ সফল হতো না। | |
| **Translate**: Liberation War would not have been successful without the leadership of Bangabandhu. | ✓ |
| 2. মীর জাফরের নেতৃত্ব ছাড়া মুক্তিযুদ্ধ সফল হতো না। | |
| **Translate**: Liberation War would not have been successful without the leadership of Mir Jafor. | ✗ |

## 5 LIMITATION & FUTURE WORK

During our analysis, we discovered that data samples containing humorous/sarcastic words, politically controversial terms, provincial vitriols, and indirect attacking words led to incorrect predictions by our fine-tuned model, BanglaBert. This may be attributed to the model being primarily trained on formal texts, necessitating updates with noisy/informal text for improved performance. Furthermore, the SentiGold training dataset may not encompass all possible witty or harsh words. Therefore, future work will involve incorporating additional examples to address ambiguous word usage and improve prediction accuracy.

## 6 CONCLUSION

Low-resource languages with small-scale data make it difficult to create task-specific models. To enrich the linguistic resources available for Bangla, particularly for the sentiment analysis task, we have introduced SentiGOLD dataset, which is ~4.5X times larger than the previous sentiment analysis resources available in Bangla as well as covers 30 domains, which is more than twice of those available previously. While developing SentiGOLD, we have incorporated a novel annotation procedure that improved the annotation quality. This procedure can be adapted to create more resources for Bangla or other South Asian languages which fall under the same language tree. We will continue working with different modeling architectures to improve the benchmark results and add more diverse complex data from more domains into the dataset for better solutions. We believe this contribution, serving as a competitive benchmark, will instigate the research community to pursue further improvements.

## 7 ACKNOWLEDGEMENT

This research work is implemented by the "Enhancement of Bangla Language in ICT through Research & Development[5]" (EBLICT) project of Bangladesh Computer Council[6] under the ICT Division, Government of Bangladesh. We want to thank them and their developing consultant team Giga Tech Limited[7] and also the testing consultant team from the Department of CSE, Bangladesh University of Engineering & Technology[8]. We would also like to thank the People's Republic of Bangladesh for financing this work.

## REFERENCES
[1] Martín Abadi, Ashish Agarwal, Paul Barham, Eugene Brevdo, Zhifeng Chen, Craig Citro, Greg S. Corrado, Andy Davis, Jeffrey Dean, Matthieu Devin, Sanjay Ghemawat, Ian Goodfellow, Andrew Harp, Geoffrey Irving, Michael Isard,

---
[5] http://eblict.gov.bd
[6] https://bcc.gov.bd
[7] https://gigatechltd.com
[8] https://buet.ac.bd/

Yangqing Jia, Rafal Jozefowicz, Lukasz Kaiser, Manjunath Kudlur, Josh Levenberg, Dandelion Mané, Rajat Monga, Sherry Moore, Derek Murray, Chris Olah, Mike Schuster, Jonathon Shlens, Benoit Steiner, Ilya Sutskever, Kunal Talwar, Paul Tucker, Vincent Vanhoucke, Vijay Vasudevan, Fernanda Viégas, Oriol Vinyals, Pete Warden, Martin Wattenberg, Martin Wicke, Yuan Yu, and Xiaoqiang Zheng. 2015. TensorFlow: Large-Scale Machine Learning on Heterogeneous Systems. https://www.tensorflow.org/ Software available from tensorflow.org.

[2] Md. Akhter-Uz-Zaman Ashik, Shahriar Shovon, and Summit Haque. 2019. Data Set For Sentiment Analysis On Bengali News Comments And Its Baseline Evaluation. In *2019 International Conference on Bangla Speech and Language Processing (ICBSLP)*. 1–5. https://doi.org/10.1109/ICBSLP47725.2019.201497

[3] Bangladesh, CPTU 2019. Central Procurement Technical Unit, Government of The People's Republic of Bangladesh. https://www.cptu.gov.bd/noa-award-details/noa-awards-1174.html.

[4] Abhik Bhattacharjee, Tahmid Hasan, Kazi Mubasshir, Md. Saiful Islam, Wasi Ahmad Uddin, Anindya Iqbal, M. Sohel Rahman, and Rifat Shahriyar. 2022. BanglaBERT: Lagnuage Model Pretraining and Benchmarks for Low-Resource Language Understanding Evaluation in Bangla. In *Findings of the North American Chapter of the Association for Computational Linguistics: NAACL 2022*. arXiv:2101.00204 https://arxiv.org/abs/2101.00204

[5] Samit Bhattacharya. 2023. Inflectional morphology synthesis for bengali noun, pronoun and verb systems. (02 2023).

[6] Nitish Ranjan Bhowmik, Mohammad Arifuzzaman, and M. Rubaiyat Hossain Mondal. 2022. Sentiment analysis on Bangla text using extended lexicon dictionary and deep learning algorithms. *Array* 13 (2022), 100123. https://doi.org/10.1016/j.array.2021.100123

[7] Washington Cunha, Vítor Mangaravite, Christian Gomes, Sérgio Canuto, Elaine Resende, Cecilia Nascimento, Felipe Viegas, Celso França, Wellington Santos Martins, Jussara M. Almeida, Thierson Rosa, Leonardo Rocha, and Marcos André Gonçalves. 2021. On the cost-effectiveness of neural and non-neural approaches and representations for text classification: A comprehensive comparative study. *Information Processing & Management* 58, 3 (2021), 102481. https://doi.org/10.1016/j.ipm.2020.102481

[8] Jianfeng Deng, Lianglun Cheng, and Zhuowei Wang. 2021. Attention-based BiLSTM fused CNN with gating mechanism model for Chinese long text classification. *Computer Speech & Language* 68 (2021), 101182.

[9] Jacob Devlin, Ming-Wei Chang, Kenton Lee, and Kristina Toutanova. 2018. BERT: Pre-training of Deep Bidirectional Transformers for Language Understanding. *CoRR* abs/1810.04805 (2018). arXiv:1810.04805 http://arxiv.org/abs/1810.04805

[10] Joseph L. Fleiss. 1971. Measuring nominal scale agreement among many raters. *Psychological Bulletin* 76 (11 1971), 378–382. Issue 5. https://doi.org/10.1037/H0031619

[11] Alec Go, Richa Bhayani, and Lei Huang. 2009. Twitter Sentiment Classification using Distant Supervision. *Processing* (2009), 1–6. http://www.stanford.edu/~alecmgo/papers/TwitterDistantSupervision09.pdf

[12] Eftekhar Hossain, Omar Sharif, Mohammed Moshiul Hoque, and Iqbal H. Sarker. 2021. SentiLSTM: A Deep Learning Approach for Sentiment Analysis of Restaurant Reviews. In *Hybrid Intelligent Systems*, Ajith Abraham, Thomas Hanne, Oscar Castillo, Niketa Gandhi, Tatiane Nogueira Rios, and Tzung-Pei Hong (Eds.). Springer International Publishing, Cham, 193–203.

[13] Minqing Hu and Bing Liu. 2004. Mining and Summarizing Customer Reviews. In *Proceedings of the Tenth ACM SIGKDD International Conference on Knowledge Discovery and Data Mining* (Seattle, WA, USA) *(KDD '04)*. Association for Computing Machinery, New York, NY, USA, 168–177. https://doi.org/10.1145/1014052.1014073

[14] ICT Divison, Government of The People's Republic of Bangladesh 22May, 2019. Request for EOI, Development of Sentiment Analysis Software in Bangla(SD-18). https://tinyurl.com/EOI-SentimentAnalysisBangla.

[15] ICT Divison, Government of The People's Republic of Bangladesh 22May, 2019. Terms of Reference, Development of Sentiment Analysis Software in Bangla(SD-18). https://drive.google.com/file/d/1KUeW-CB1AaJpvnpc9iTTB3hMq0hBZ3hO/view?usp=sharing.

[16] Md Shafiqul Islalm, Md Moklesur Rahman, Md Hafizur Rahman, Md Arifuzzaman, Roberto Sassi, and Md Aktaruzzaman. 2019. Recognition bangla sign language using convolutional neural network. In *2019 international conference on innovation and intelligence for informatics, computing, and technologies (3ICT)*. IEEE, 1–6.

[17] Khondoker Ittehadul Islam, Md Saiful Islam, and Md Ruhul Amin. 2020. Sentiment analysis in Bengali via transfer learning using multi-lingual BERT. *ICCIT 2020 - 23rd International Conference on Computer and Information Technology, Proceedings*. https://doi.org/10.1109/ICCIT51783.2020.9392653

[18] Khondoker Ittehadul Islam, Md Saiful Islam, and Md Ruhul Amin. 2020. Sentiment analysis in Bengali via transfer learning using multi-lingual BERT. In *2020 23rd International Conference on Computer and Information Technology (ICCIT)*. 1–5. https://doi.org/10.1109/ICCIT51783.2020.9392653

[19] Khondoker Ittehadul Islam, Sudipta Kar, Md Saiful Islam, and Mohammad Ruhul Amin. 2021. SentNoB: A Dataset for Analysing Sentiment on Noisy Bangla Texts. In *Findings of the Association for Computational Linguistics: EMNLP 2021*. Association for Computational Linguistics, Punta Cana, Dominican Republic, 3265–3271. https://doi.org/10.18653/v1/2021.findings-emnlp.278

[20] Paul Jaccard. 1908. Nouvelles Recherches Sur la Distribution Florale. *Bulletin de la Societe Vaudoise des Sciences Naturelles* 44 (01 1908), 223–70. https://doi.org/10.5169/seals-268384

[21] Mst Eshita Khatun and Tapasy Rabeya. 2022. A Machine Learning Approach for Sentiment Analysis of Book Reviews in Bangla Language. *2022 6th International Conference on Trends in Electronics and Informatics, ICOEI 2022 - Proceedings*, 1178–1182. https://doi.org/10.1109/ICOEI53556.2022.9776752

[22] Dimitrios Kouzis-Loukas. 2016. *Learning Scrapy*. Packt Publishing Ltd.

[23] Matt J. Kusner, Yu Sun, Nicholas I. Kolkin, and Kilian Q. Weinberger. 2015. From Word Embeddings to Document Distances. In *Proceedings of the 32nd International Conference on International Conference on Machine Learning - Volume 37* (Lille, France) *(ICML'15)*. JMLR.org, 957–966.

[24] Quoc Le and Tomas Mikolov. 2014. Distributed Representations of Sentences and Documents. In *Proceedings of the 31st International Conference on International Conference on Machine Learning - Volume 32* (Beijing, China) *(ICML'14)*. JMLR.org, II–1188–II–1196.

[25] Steven Loria. 2018. textblob Documentation. *Release 0.15* 2 (2018).

[26] Andrew L. Maas, Raymond E. Daly, Peter T. Pham, Dan Huang, Andrew Y. Ng, and Christopher Potts. 2011. Learning Word Vectors for Sentiment Analysis. In *Proceedings of the 49th Annual Meeting of the Association for Computational Linguistics: Human Language Technologies*. Association for Computational Linguistics, Portland, Oregon, USA, 142–150. https://aclanthology.org/P11-1015

[27] Tomas Mikolov, Kai Chen, Greg Corrado, and Jeffrey Dean. 2013. Efficient Estimation of Word Representations in Vector Space. https://doi.org/10.48550/ARXIV.1301.3781

[28] Md Saddam Hossain Mukta, Md Adnanul Islam, Faisal Ahamed Khan, Afjal Hossain, Shuvanon Razik, Shazzad Hossain, and Jalal Mahmud. 2021. A Comprehensive Guideline for Bengali Sentiment Annotation. *Transactions on Asian and Low-Resource Language Information Processing* 21, 2 (2021), 1–19.

[29] Forhad An Naim. 2021. Bangla Aspect-Based Sentiment Analysis Based on Corresponding Term Extraction. *undefined* (2 2021), 65–69. https://doi.org/10.1109/ICICT4SD50815.2021.9396970

[30] Rebecca Passonneau, Nizar Habash, and Owen Rambow. 2006. Inter-annotator Agreement on a Multilingual Semantic Annotation Task. In *Proceedings of the Fifth International Conference on Language Resources and Evaluation (LREC'06)*. European Language Resources Association (ELRA), Genoa, Italy. http://www.lrec-conf.org/proceedings/lrec2006/pdf/634_pdf.pdf

[31] Adam Paszke, Sam Gross, Francisco Massa, Adam Lerer, James Bradbury, Gregory Chanan, Trevor Killeen, Zeming Lin, Natalia Gimelshein, Luca Antiga, Alban Desmaison, Andreas Köpf, Edward Yang, Zachary DeVito, Martin Raison, Alykhan Tejani, Sasank Chilamkurthy, Benoit Steiner, Lu Fang, Junjie Bai, and Soumith Chintala. 2019. PyTorch: An Imperative Style, High-Performance Deep Learning Library.. In *NeurIPS*, Hanna M. Wallach, Hugo Larochelle, Alina Beygelzimer, Florence d'Alché Buc, Emily B. Fox, and Roman Garnett (Eds.). 8024–8035. http://dblp.uni-trier.de/db/conf/nips/nips2019.html#PaszkeGMLBCKLGA19

[32] Marco Tulio Ribeiro, Tongshuang Wu, Carlos Guestrin, and Sameer Singh. 2020. Beyond Accuracy: Behavioral Testing of NLP Models with CheckList. In *Proceedings of the 58th Annual Meeting of the Association for Computational Linguistics*. Association for Computational Linguistics, Online, 4902–4912. https://doi.org/10.18653/v1/2020.acl-main.442

[33] Salim Sazzed. 2020. Cross-lingual Sentiment Analysis in Bengali Utilizing A New Benchmark Corpus. (2020). https://github.com/sazzadcsedu/BN-Dataset.git

[34] Salim Sazzed. 2020. Cross-lingual sentiment classification in low-resource Bengali language. In *Proceedings of the Sixth Workshop on Noisy User-generated Text (W-NUT 2020)*. Association for Computational Linguistics, Online, 50–60. https://doi.org/10.18653/v1/2020.wnut-1.8

[35] Richard Socher, Alex Perelygin, Jean Wu, Jason Chuang, Christopher D. Manning, Andrew Ng, and Christopher Potts. 2013. Recursive Deep Models for Semantic Compositionality Over a Sentiment Treebank. In *Proceedings of the 2013 Conference on Empirical Methods in Natural Language Processing*. Association for Computational Linguistics, Seattle, Washington, USA, 1631–1642. https://aclanthology.org/D13-1170

[36] Md Ferdous Wahid, Md Jahid Hasan, and Md Shahin Alom. 2019. Cricket Sentiment Analysis from Bangla Text Using Recurrent Neural Network with Long Short Term Memory Model. *2019 International Conference on Bangla Speech and Language Processing, ICBSLP 2019* (9 2019). https://doi.org/10.1109/ICBSLP47725.2019.201500

[37] Thomas Wolf, Lysandre Debut, Victor Sanh, Julien Chaumond, Clement Delangue, Anthony Moi, Pierric Cistac, Tim Rault, Remi Louf, Morgan Funtowicz, Joe Davison, Sam Shleifer, Patrick von Platen, Clara Ma, Yacine Jernite, Julien Plu, Canwen Xu, Teven Le Scao, Sylvain Gugger, Mariama Drame, Quentin

Lhoest, and Alexander Rush. 2020. Transformers: State-of-the-Art Natural Language Processing. In *Proceedings of the 2020 Conference on Empirical Methods in Natural Language Processing: System Demonstrations*. Association for Computational Linguistics, Online, 38–45. https://doi.org/10.18653/v1/2020.emnlp-demos.6

[38] Zichao Yang, Diyi Yang, Chris Dyer, Xiaodong He, Alex Smola, and Eduard Hovy. 2016. Hierarchical attention networks for document classification. In *Proceedings of the 2016 conference of the North American chapter of the association for computational linguistics: human language technologies*. 1480–1489.

[39] Jerrold H Zar. 1999. *Biostatistical analysis*. Pearson Education India.

[40] Lei Zhang, Shuai Wang, and Bing Liu. 2018. Deep Learning for Sentiment Analysis : A Survey. *CoRR* abs/1801.07883 (2018). arXiv:1801.07883 http://arxiv.org/abs/1801.07883

## A APPENDIX

### A.1 Data Source

The data was gathered from the following website. We divide this sources into five categories:

**Social Media:** youtube.com, facebook.com
**Product Review:** daraz.com, rokomari.com, evaly.com
**News:** prothomalo.com, jugantor.com, bd-pratidin.com, kalerkantho.com, ittefaq.com, dailyinqilab.com, samakal.com, dailynayadiganta.com, manobkantha.com, jaijaidinbd.com, bd-journal.com
**Bangla Blogs:** amrabondhu.com, cadetcollegeblog.com, mukto-mona.com, choturmatrik.com, bishorgo.com, sachalayatan.com, blogspot.com, neurogenbd.com
**Tech Blogs:** banglatech24,com, techtunes.com, somewhereinblog.net

### A.2 Source Distribution

**Figure 12** represents the source distribution of our dataset.

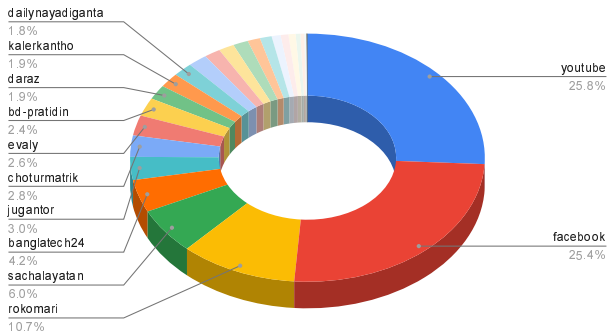

**Figure 12: Illustration of our proposed dataset's source distribution. Youtube and facebook domainates the distribution as they contain more subjective sentences which also helped in balancing the class distribution.**

### A.3 Data Deduplication Procedure

For data deduplication we utilize Jaccard Similarity [20] score and Word Mover Distance (WMD) [23]. The threshold values were set at 0.45 for Jaccard similarity and 0.4 for WMD. If Jaccard similarity was greater than 0.45 and WMD was less than 0.4 between two

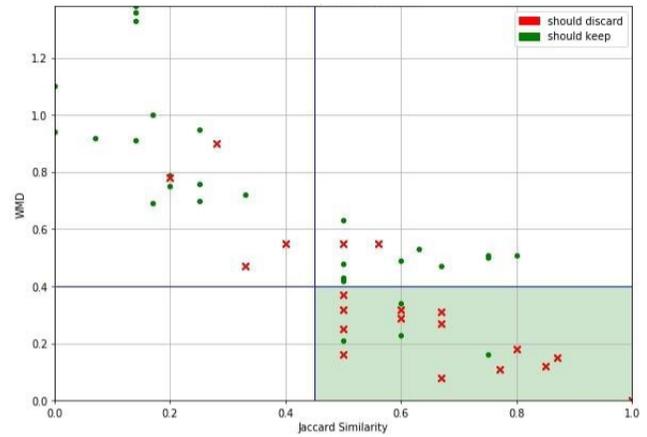

**Figure 13: The figure illustrates 50 sentence pairs that have been annotated as either "keep both" or "discard one" plotted on a graph where the x-axis represents their Jaccard similarity, and the y-axis represents their Word Mover Distance (WMD). The graph shows that 85% of the pairs that should be kept have a Jaccard similarity <0.45 and a WMD >0.4.**

sentences, then one of them were eliminated otherwise, both were retained.

In order to establish the threshold values, we tagged 50 pairs of sentences as either "keep both" or "discard one". We then calculated the Jaccard and WMD scores for each pair and plotted them on a two-dimensional graph (see **Figure 13**). We infer the threshold values from the shaded region.

### A.4 Statistical Significance Test

We have done 10-fold cross-validation using the SentiGOLD dataset across our 20 models. The results from **Table 8** shows that BanglaBert gave the best performance in every fold.

As we have multiple methods, we use the non-parametric Friedman test to determine if there are any significant differences between the median values of the methods. We use the post-hoc Nemenyi test to infer which differences are significant. We report the median (MD), the median absolute deviation (MAD) and the mean rank (MR), among all methods over the samples in **Table 9**. Differences between methods are significant, if the difference of the mean rank is greater than the critical distance CD=9.376 of the Nemenyi test.

**Figure 14** shows the critical difference diagram summarizing the outcome of friedman test followed by the post-hoc Nemenyi test, where lines uniting two or more methods indicate that there are no statistical differences among them.

According to the above analysis, we come to the conclusion that the BanglaBert model provides us with consistent Macro-F1 across all the 10-folds among the 20 models we selected. In addition to that, we observe that the combination of features like unigram, bigram, trigram and C2 to C5 character n-grams can also provide consistent performance though their Macro-F1 is lower than BanglaBert with a higher margin.

Table 8: We show the results (Macro-F1) for 10-fold cross-validation using the SentiGOLD dataset for 20 different models.

| Model Name | Fold-1 (Macro F1) | Fold-2 (Macro F1) | Fold-3 (Macro F1) | Fold-4 (Macro F1) | Fold-5 (Macro F1) | Fold-6 (Macro F1) | Fold-7 (Macro F1) | Fold-8 (Macro F1) | Fold-9 (Macro F1) | Fold-10 (Macro F1) |
|---|---|---|---|---|---|---|---|---|---|---|
| Unigram | 0.47 | 0.47 | 0.47 | 0.47 | 0.47 | 0.47 | 0.48 | 0.47 | 0.47 | 0.47 |
| Bigram | 0.4 | 0.39 | 0.38 | 0.39 | 0.39 | 0.4 | 0.4 | 0.38 | 0.39 | 0.4 |
| Trigram | 0.24 | 0.26 | 0.25 | 0.25 | 0.24 | 0.25 | 0.25 | 0.24 | 0.25 | 0.25 |
| U+B | 0.5 | 0.49 | 0.49 | 0.49 | 0.5 | 0.5 | 0.5 | 0.49 | 0.49 | 0.49 |
| B+T | 0.4 | 0.39 | 0.39 | 0.39 | 0.4 | 0.4 | 0.41 | 0.38 | 0.39 | 0.4 |
| U+B+T | 0.5 | 0.49 | 0.49 | 0.49 | 0.5 | 0.49 | 0.5 | 0.49 | 0.49 | 0.49 |
| Char 2-gram | 0.44 | 0.43 | 0.43 | 0.43 | 0.44 | 0.44 | 0.44 | 0.43 | 0.43 | 0.44 |
| Char 3-gram | 0.48 | 0.48 | 0.48 | 0.48 | 0.49 | 0.48 | 0.49 | 0.48 | 0.47 | 0.48 |
| Char 4-gram | 0.49 | 0.49 | 0.49 | 0.48 | 0.49 | 0.48 | 0.49 | 0.49 | 0.49 | 0.48 |
| Char 5-gram | 0.5 | 0.5 | 0.49 | 0.49 | 0.5 | 0.48 | 0.5 | 0.49 | 0.48 | 0.49 |
| C2+C3 | 0.48 | 0.49 | 0.48 | 0.48 | 0.49 | 0.48 | 0.49 | 0.48 | 0.48 | 0.49 |
| C3+C4 | 0.5 | 0.5 | 0.5 | 0.49 | 0.5 | 0.49 | 0.5 | 0.49 | 0.49 | 0.49 |
| C4+C5 | 0.49 | 0.5 | 0.5 | 0.49 | 0.5 | 0.49 | 0.5 | 0.49 | 0.49 | 0.5 |
| C2+C3+C4 | 0.5 | 0.5 | 0.5 | 0.49 | 0.5 | 0.49 | 0.5 | 0.5 | 0.49 | 0.49 |
| C3+C4+C5 | 0.5 | 0.51 | 0.5 | 0.49 | 0.51 | 0.5 | 0.51 | 0.5 | 0.5 | 0.5 |
| C2+C3+C4+C5 | 0.5 | 0.51 | 0.5 | 0.5 | 0.52 | 0.5 | 0.51 | 0.51 | 0.5 | 0.5 |
| U+B+C3+C4+C5 | 0.49 | 0.49 | 0.49 | 0.48 | 0.5 | 0.49 | 0.49 | 0.49 | 0.49 | 0.48 |
| U+B+C2+C3+C4+C5 | 0.49 | 0.49 | 0.49 | 0.49 | 0.5 | 0.49 | 0.49 | 0.49 | 0.49 | 0.48 |
| U+B+T+C2+C3+C4+C5 | 0.5 | 0.5 | 0.49 | 0.5 | 0.51 | 0.5 | 0.5 | 0.49 | 0.5 | 0.49 |
| **BanglaBERT** | **0.64** | **0.65** | **0.64** | **0.64** | **0.65** | **0.65** | **0.65** | **0.64** | **0.65** | **0.64** |

Table 9: Results of post-hoc Nemenyi test to report the Mean Rank, Median of Macro-F1, the Median Absolute Deviation (MAD) and the Confidence Interval of the Median of Macro-F1.

| Model Name | Mean Rank | Median | Median Absolute Deviation | Confidence Interval |
|---|---|---|---|---|
| Trigram | 20 | 0.25 | 0 | [0.240, 0.260] |
| Bigram | 18.65 | 0.39 | 0.01 | [0.380, 0.400] |
| B+T | 18.35 | 0.395 | 0.005 | [0.380, 0.410] |
| Char 2-gram | 17 | 0.435 | 0.005 | [0.430, 0.440] |
| Unigram | 15.95 | 0.47 | 0 | [0.470, 0.480] |
| Char 3-gram | 14.15 | 0.48 | 0 | [0.470, 0.490] |
| C2+C3 | 13.05 | 0.48 | 0 | [0.480, 0.490] |
| Char 4-gram | 11.8 | 0.49 | 0 | [0.480, 0.490] |
| U+B+C3+C4+C5 | 10.75 | 0.49 | 0 | [0.480, 0.500] |
| U+B+C2+C3+C4+C5 | 10.15 | 0.49 | 0 | [0.480, 0.500] |
| Char 5-gram | 8.85 | 0.49 | 0.01 | [0.480, 0.500] |
| U+B+T | 8.4 | 0.49 | 0 | [0.490, 0.500] |
| U+B | 7.9 | 0.49 | 0 | [0.490, 0.500] |
| C4+C5 | 7.35 | 0.495 | 0.005 | [0.490, 0.500] |
| C3+C4 | 7.25 | 0.495 | 0.005 | [0.490, 0.500] |
| C2+C3+C4 | 6.7 | 0.5 | 0 | [0.490, 0.500] |
| U+B+T+C2+C3+C4+C5 | 5.8 | 0.5 | 0 | [0.490, 0.510] |
| C3+C4+C5 | 3.85 | 0.5 | 0 | [0.490, 0.510] |
| C2+C3+C4+C5 | 3.05 | 0.5 | 0 | [0.500, 0.520] |
| BanglaBERT | 1 | 0.645 | 0.005 | [0.640, 0.650] |

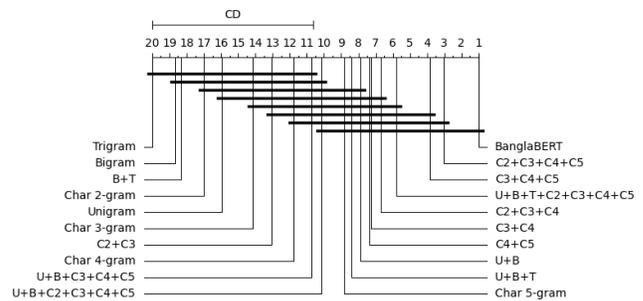

Figure 14: The critical differences diagram, obtained from the Friedman test followed by the post-hoc Nemenyi test, with a significance level of 95%. Values to the right indicate a better average rank performance.

## A.5 Data Licensing & Legality

The SentiGOLD project adhered to the highest ethical standards and fair data usage policies of social media platforms. The project, funded by Bangladesh Computer Council (BCC) under the EBLICT[9] program, limited dataset usage to non-commercial and academic research. Data collection followed the fair usage policies of Facebook[10] and YouTube[11], using standard API calls and random sampling from public pages and comments. Anonymization and de-identification steps were taken to protect privacy, and sentiment labels were added to the collected data with an IAA > 0.80. Personal mentions were anonymized, ensuring anonymity and de-identification in accordance with EBLICT rules.

## A.6 Public Release of the Dataset

The release of the SentiGOLD dataset will adhere to a data sharing protocol governed by EBLICT policy. To request access to the dataset, researchers must submit a data request form[12] followed by an End-User License Agreement (EULA) with EBLICT.

---

[9] http://eblict.gov.bd
[10] https://www.facebook.com/help/1020633957973118
[11] https://support.google.com/youtube/answer/9783148?hl=en
[12] https://forms.gle/vHtTmbWnajMuLHY17